\documentclass[10pt,twocolumn,letterpaper]{article}

\usepackage[pagenumbers]{wacv}

\usepackage[utf8]{inputenc}
\usepackage[T1]{fontenc}
\usepackage{amsfonts}
\usepackage{nicefrac}
\usepackage{microtype}
\usepackage{bm}
\usepackage{bbding}
\usepackage{colortbl}
\usepackage{multirow}

\definecolor{wacvblue}{rgb}{0.21,0.49,0.74}
\usepackage[pagebackref,breaklinks,colorlinks,allcolors=wacvblue]{hyperref}

\def\wacvPaperID{440}
\def\confName{WACV}
\def\confYear{2027}

\title{Learning Motion-Expressive 4D Gaussian Representations from Monocular Videos with Flow Splatting}

\author{
\bfseries
Jinzhao Li,
Miao Liu \\[1em]
Tsinghua University\\[1em]
\texttt{lijinzha22@mails.tsinghua.edu.cn} \quad \texttt{miaoliu@mail.tsinghua.edu.cn}
}

\newlength{\wacvteaserheight}
\newcommand{\wacvteaser}{%
\par\noindent
\begin{minipage}{\textwidth}
    \captionsetup{type=figure}
    \setlength{\wacvteaserheight}{0.35\linewidth}
    \centering
    \begin{subfigure}[t]{0.61\textwidth}
        \centering
        \includegraphics[height=\wacvteaserheight,keepaspectratio]{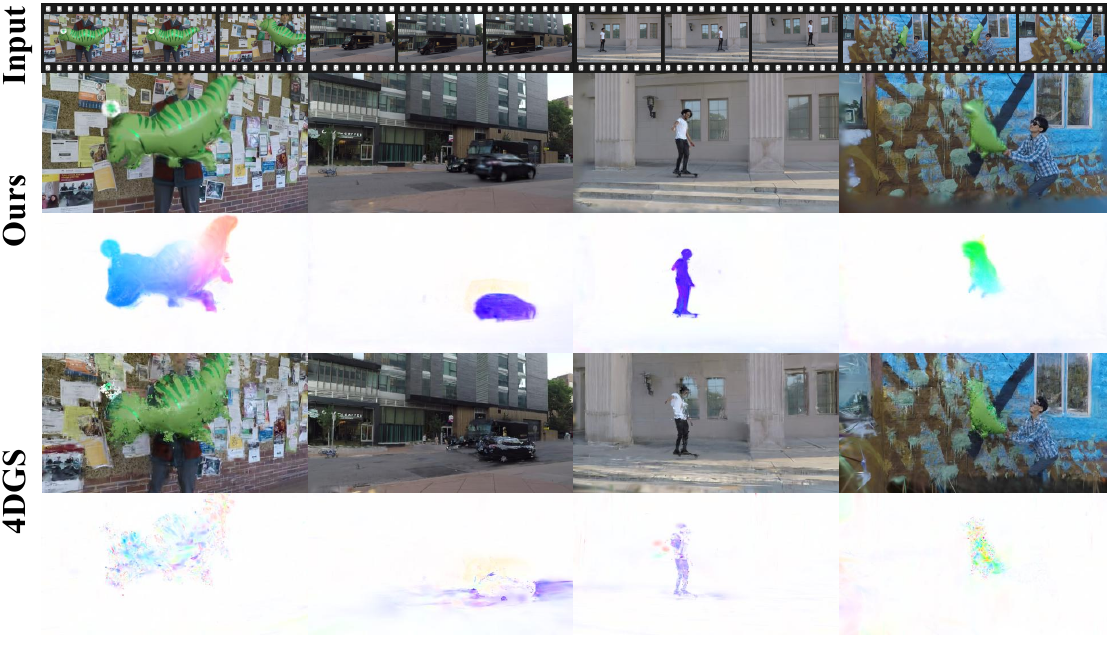}
        \caption{Rendering optical flow}
        \label{fig:teaser_1}
    \end{subfigure}
    \hfill
    \begin{subfigure}[t]{0.35\textwidth}
        \centering
        \includegraphics[height=\wacvteaserheight,keepaspectratio]{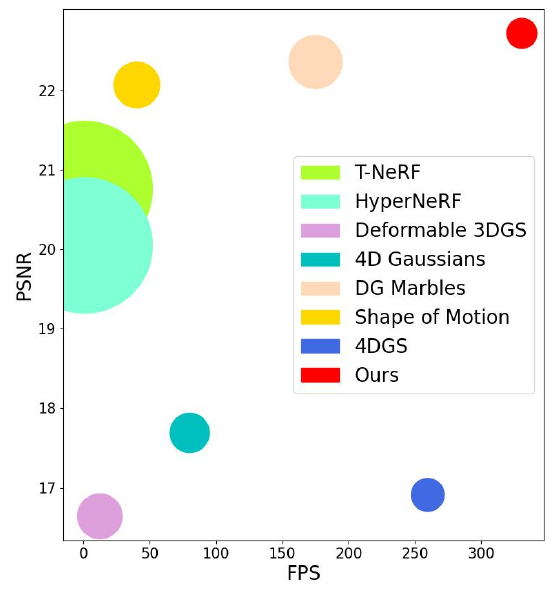}
        \caption{Quantitative results}
        \label{fig:teaser_2}
    \end{subfigure}
    \caption{\textbf{Comparison of previous methods and ours.} (a) We visualize the rendering results of color and optical flow for our baseline~\cite{Real-time2024ICLR} and Flow Splatting. (b) We report PSNR and the rendering speed on the NVIDIA dataset~\cite{NVIDIA2020CVPR} for multiple methods. The size of circle represents training time.}
    \label{fig:teaser}
\end{minipage}
\par\vspace{0.5em}%
}

\makeatletter
\apptocmd{\@maketitle}{\wacvteaser}{}{%
    \PackageError{main}{Unable to append the teaser to the title block}{}
}
\makeatother

\begin{document}

\maketitle

\begin{abstract}
\noindent
Reconstructing dynamic 3D scenes from monocular videos remains challenging because limited observations provide insufficient motion constraints.
Conventional 4D Gaussian representations model spatiotemporal appearance efficiently, but the restricted temporal evolution of each primitive makes complex nonlinear dynamics difficult to represent compactly.
We propose a motion-expressive 4D Gaussian representation with time-varying means and covariances.
By analytically differentiating Gaussian trajectories, we obtain Gaussian velocities and introduce Flow Splatting to render optical flow while accounting for camera motion, enabling dense motion supervision.
This formulation jointly models scene appearance and motion within a unified set of 4D Gaussian primitives.
We further introduce an initialization strategy based on the sampling theorem and a velocity consistency regularizer to stabilize optimization.
Experiments on DAVIS and NVIDIA Dynamic Scenes demonstrate a favorable trade-off between reconstruction quality and efficiency under challenging monocular settings.
\end{abstract}

\section{Introduction}\label{sec:introduction}

Reconstructing scenes from 2D images has been a long-standing goal in computer vision due to its widespread applications, including virtual reality~\cite{Vr-gs2024Siggraph}, robotics~\cite{Robot2022RAL}, and autonomous driving~\cite{Drivinggaussian2024CVPR}. Recent advances in visual scene understanding~\cite{chen2026learning,chen2026snapbench,li2026ipibench,Li_2026_CVPR} further highlight the growing importance of effective visual representations across diverse applications.
Remarkable progress has been made using neural implicit representations~\cite{SRN2019NIPS,NeRF2020ECCV,LFN2021NIPS}, but these methods often incur substantial training and rendering costs~\cite{EfficientNeRF2022CVPR,FastNeRF2021ICCV,InstantNGP2022TOG,KiloNeRF2021ICCV,MipNeRF2021ICCV,NSVF2020NIPS,PlenOctree2021CVPR,Plenoxels2022CVPR}.
Recent advances in this area have been largely driven by 3D Gaussian Splatting (3DGS)~\cite{3DGS2023ToG}, which provides an explicit scene representation with real-time rendering performance.
Benefiting from rasterization-based rendering, 3DGS avoids dense point queries in scene space and achieves both high efficiency and rendering quality.
However, vanilla 3DGS mainly focuses on static scene reconstruction.

To model dynamic scenes, some methods learn deformation fields~\cite{Deformable3DGS2024CVPR,4DGS2024CVPR,Gaufre2025WACV,MoDGS2025ICLR} to deform 3D Gaussians through neural networks, while others explicitly model Gaussian motion by optimizing trajectories~\cite{Dynamic3DGS20243DV,Shapeofmotion2024arXiv,Mosca2024arXiv}.
Although these approaches can represent flexible motion, they often require lengthy optimization and may overfit under monocular settings.
More recently, several methods~\cite{4d-rotor2024Siggraph,Real-time2024ICLR} treat space and time jointly and directly represent dynamic scenes with 4D Gaussians.
Such representations can naturally handle the appearance and disappearance of scene content.
However, the motion represented by each conventional 4D Gaussian is limited, making complex nonlinear dynamics difficult to model with a compact set of primitives.
Moreover, the lack of dense motion supervision under insufficient viewpoints weakens temporal constraints~\cite{Motion-aware2024TCSVT}, which can lead to temporally inconsistent motion, visual overfitting, and redundant scene modeling.

To tackle these challenges, we propose a motion-expressive 4D Gaussian representation together with Flow Splatting for dynamic scene reconstruction from monocular videos.
We extend conventional 4D Gaussians~\cite{4d-rotor2024Siggraph,Real-time2024ICLR} with time-varying means and covariances to represent complex nonlinear dynamics.
Based on this representation, we analytically derive a velocity field from the temporal derivative of the Gaussian trajectories.
We then extend differentiable Gaussian splatting to render the velocity field in image space while accounting for camera motion, enabling dense supervision from optical flow.
Furthermore, we introduce an initialization strategy based on the sampling theorem and a velocity consistency regularizer to stabilize optimization and improve motion reconstruction.
In this manner, appearance and motion are jointly modeled within a unified 4D Gaussian representation.
The proposed representation can mitigate degeneration into per-frame reconstruction and reduce redundant primitives, while Flow Splatting provides dense motion constraints that improve temporal consistency.

We conduct extensive experiments to demonstrate the effectiveness and efficiency of our method.
As shown in Figure~\ref{fig:teaser_1}, our method reconstructs scene appearance and dynamic motion more faithfully than the conventional 4DGS baseline~\cite{Real-time2024ICLR}.
The quantitative results in Figure~\ref{fig:teaser_2} further demonstrate a favorable trade-off between rendering quality and efficiency.
Our main contributions can be summarized as follows:
\begin{itemize}[leftmargin=*]
    \item We propose a motion-expressive 4D Gaussian representation with time-varying means and covariances to model complex nonlinear dynamics using a compact set of primitives.
    
    \item We analytically derive Gaussian velocities and introduce Flow Splatting, which accounts for camera motion to render optical flow for dense motion supervision. We further develop an initialization strategy based on the sampling theorem and a velocity consistency regularizer for stable optimization.
    
    \item Extensive experiments on the DAVIS~\cite{Davis2017arXiv} and NVIDIA~\cite{NVIDIA2020CVPR} Dynamic Scenes datasets demonstrate that our method outperforms previous methods in terms of visual quality and efficiency.
\end{itemize}

\section{Related Works}\label{sec:related works}

\subsection{Static Novel View Synthesis}
Early work on novel view synthesis often relies on densely captured views for scene reconstruction, while neural implicit representations~\cite{Deepsdf2019CVPR, SRN2019NIPS, MVNS2020NIPS, OccNet2019CVPR} have significantly advanced the representation of 3D scenes from multi-view images.
Neural Radiance Fields (NeRF)~\cite{NeRF2020ECCV} pioneered the use of fully connected neural networks to synthesize images from novel viewpoints.
Subsequent works have addressed its limitations by improving training and rendering efficiency~\cite{EfficientNeRF2022CVPR, FastNeRF2021ICCV,InstantNGP2022TOG, KiloNeRF2021ICCV, NSVF2020NIPS,PlenOctree2021CVPR, Plenoxels2022CVPR}, extending reconstruction to large urban scenes~\cite{MegaNeRF2022CVPR, BlockNeRF2022CVPR, GridNeRF2023CVPR, BungeeNeRF2022ECCV}, and enabling reconstruction from sparse input views~\cite{niemeyer2022regnerf,truong2023sparf,wynn2023diffusionerf,xu2023murf}.

More recently, 3D Gaussian Splatting (3DGS)~\cite{3DGS2023ToG} has attracted significant attention for novel view synthesis.
Unlike the expensive volume sampling used in NeRF, 3DGS employs an efficient rasterization-based splatting process to render novel views from a set of explicit 3D Gaussian primitives.
Subsequent works have improved rendering quality and photorealism~\cite{yan2023multi, gao2023relightable, jiang2023gaussianshader, liang2023gs}, reduced memory consumption~\cite{lu2023scaffold, navaneet2023compact3d, girish2023eagles, fan2023lightgaussian, katsumata2023efficient}, and enhanced feed-forward generalization~\cite{pixelSplat2023arXiv, MVSplat2024arXiv, SplatterImage2023arXiv, GGN2024NIPS}.

\subsection{Dynamic Novel View Synthesis}

Dynamic scene reconstruction and novel view synthesis have also been long-standing problems.
One line of work extends NeRF by incorporating time as an additional input dimension~\cite{NRFlow2021CS, DynNeRF2021ICCV, NeuralSceneFlow2021CVPR, NeuralVolumes2019arXiv, Space-time2021CVPR}.
Other methods construct a canonical space and map it to individual time steps using scene flow or motion fields~\cite{Nerfies2021ICCV, D-NeRF2021CVPR}.
For example, DyNeRF~\cite{Neural3d2022CVPR} represents continuous space-time radiance fields using a series of temporal latent embeddings, while Nerfies~\cite{Nerfies2021ICCV} and HyperNeRF~\cite{Hypernerf2021arXiv} model scene dynamics through deformation fields that map observations to a canonical space.

Another line of work models dynamic scenes with 3DGS.
Some methods~\cite{Deformable3DGS2024CVPR, 4DGS2024CVPR, Gaufre2025WACV, MoDGS2025ICLR} employ time-conditioned deformation networks.
For example, Deformable 3DGS~\cite{Deformable3DGS2024CVPR} proposes a deformable version of 3DGS by introducing a deformation MLP network to model the 3D flows, while 4D Gaussians~\cite{4DGS2024CVPR} uses a more efficient Hexplane representations~\cite{Hexplane2023CVPR}.
Other works~\cite{Dynamic3DGS20243DV, Shapeofmotion2024arXiv, Mosca2024arXiv} explicitly learn 3D Gaussian trajectories over time by sequentially optimizing offsets over frames.
However, these methods always suffer from long training time or low inference speed.
Meanwhile, several methods~\cite{4d-rotor2024Siggraph, Real-time2024ICLR} directly extend 3D Gaussians into 4D space by jointly modeling space and time.
However, the temporal evolution represented by each conventional 4D Gaussian remains restricted, making complex nonlinear motion difficult to capture with a compact set of primitives.
This limitation is further amplified in monocular settings, where insufficient motion constraints can lead to temporally fragmented and redundant scene representations.

\section{Methods}\label{sec:methods}

\begin{figure*}[t]
    \centering
    \includegraphics[width=\linewidth]{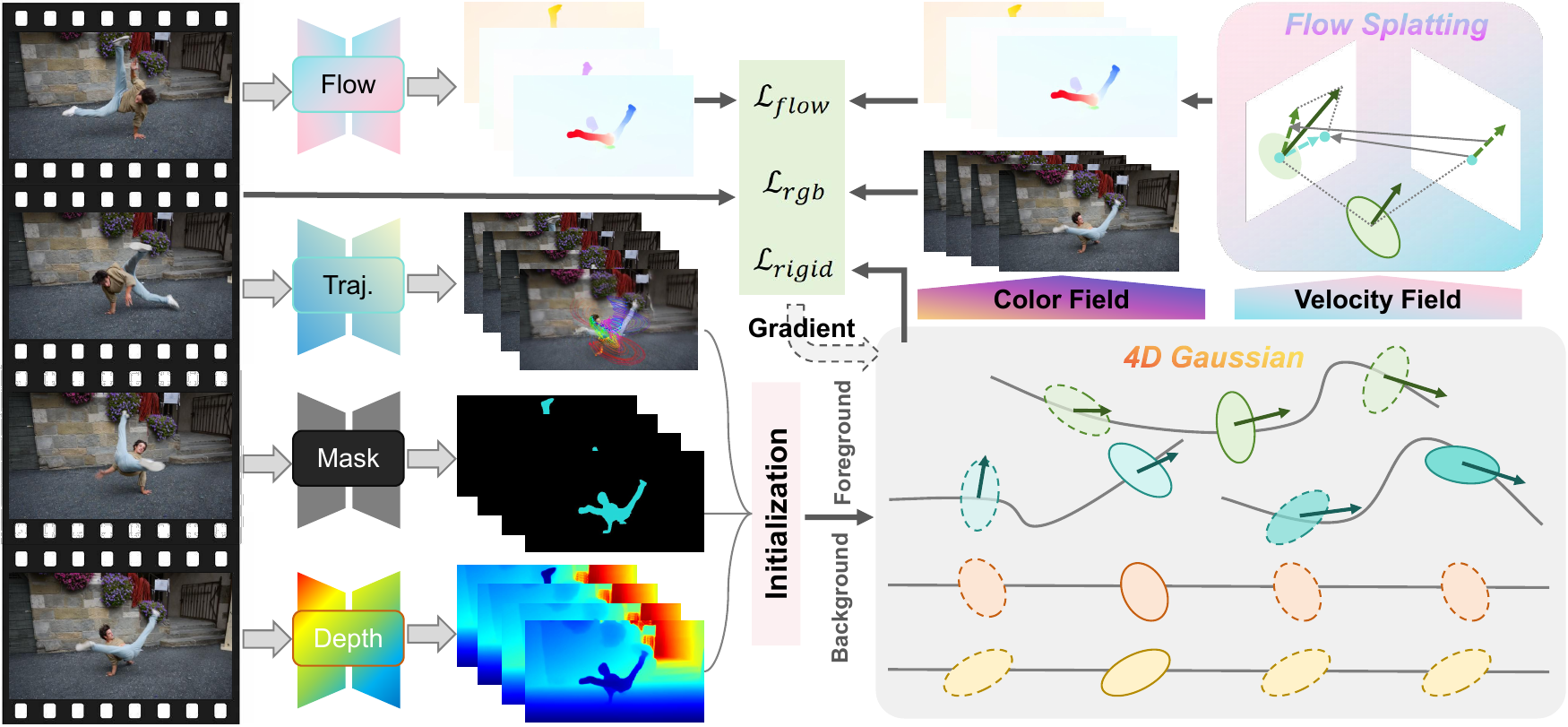}
    \caption{\textbf{Overview of our framework.}
    We first leverage off-the-shelf models to predict optical flow, depth maps, masks, and trajectories from the input monocular video.
    Then, we initialize the foreground and background separately using these data-driven priors.
    We extend conventional 4D Gaussians with time-varying means and covariances to represent complex nonlinear motion.
    Based on this representation, we derive a velocity field and introduce Flow Splatting to render optical flow while accounting for camera motion.
    We use both color and velocity losses for supervision.}
    \label{fig:pipeline}
\end{figure*}

\subsection{Preliminaries}

\noindent\textbf{Representation of 3D Gaussians.}
3DGS~\cite{3DGS2023ToG} represents a scene as a set of 3D Gaussian primitives, including a center position $\bm{\mu}\in\mathbb{R}^{3}$, a covariance matrix $\Sigma\in\mathbb{R}^{3\times 3}$, an opacity $o\in[0,1)$, and spherical harmonics coefficients $\bm{c}\in\mathbb{R}^{k}$.
The Gaussian function can be formulated as
\begin{equation}
    G(\bm{x})
    =
    e^{-\frac{1}{2}
    (\bm{x}-\bm{\mu})^\top
    \Sigma^{-1}
    (\bm{x}-\bm{\mu})}.
    \label{eq: Gaussian function}
\end{equation}
Here, $\Sigma=RSS^\top R^\top$, where $S$ is the scaling matrix and $R$ is the rotation matrix.
For every pixel, the color is rendered by a set of Gaussians sorted in depth order:
\begin{equation}
    C
    =
    \sum_{i\in N}
    \bm{c}_{i}\alpha_{i}
    \prod_{j=1}^{i-1}(1-\alpha_{j}).
    \label{eq:color rendering}
\end{equation}

\noindent\textbf{Representation of 4D Gaussians.}
Analogous to 3D Gaussians, a 4D Gaussian~\cite{4d-rotor2024Siggraph, Real-time2024ICLR} can be expressed with a 4D center position
$\bm{\mu}_{\text{4D}}=(\mu_x,\mu_y,\mu_z,\mu_t)^{\top}$
and a 4D covariance matrix $\Sigma_{\text{4D}}$.
Let $\bm{z}=(\bm{x}^{\top},t)^{\top}$ denote a point in 4D space.
The 4D Gaussian is formulated as
\begin{equation}
\begin{aligned}
    G_{\text{4D}}(\bm{z})
    =
    \exp\Big[
        -\frac{1}{2}
        &(\bm{z}-\bm{\mu}_{\text{4D}})^{\top} \\
        &\Sigma_{\text{4D}}^{-1}
        (\bm{z}-\bm{\mu}_{\text{4D}})
    \Big].
\end{aligned}
\end{equation}
The covariance $\Sigma_{\text{4D}}$ can be further factorized into the 4D scaling matrix $S_{\text{4D}}$ and the 4D rotation matrix $R_{\text{4D}}$ as
\begin{equation}
    \Sigma_{\text{4D}}
    =
    R_{\text{4D}}S_{\text{4D}}
    S_{\text{4D}}^{\top}R_{\text{4D}}^{\top}.
\end{equation}
Since $\Sigma_{\text{4D}}$ is a symmetric matrix, we write
\begin{equation}
    \Sigma_{\text{4D}}
    =
    \begin{pmatrix}
        U & V \\
        V^{\top} & W
    \end{pmatrix},
\end{equation}
where $U$ is a $3\times3$ matrix.

The projected 3D Gaussian at time $t$ is obtained as
\begin{equation}
\begin{aligned}
    G(\bm{x},t)
    &=
    p(t)p(\bm{x}\vert t) \\
    &=
    e^{-\frac{1}{2}\lambda(t-\mu_t)^2} \\
    &\quad
    e^{-\frac{1}{2}
    (\bm{x}-\bm{\mu}(t))^{\top}
    \Sigma_{\text{3D}}^{-1}
    (\bm{x}-\bm{\mu}(t))}.
\end{aligned}
\label{eq:projected 3D gaussian}
\end{equation}
Here,
$\lambda=W^{-1}$,
$\Sigma_{\text{3D}}=U-{VV^{\top}}/{W}$, and
\begin{equation}
    \bm{\mu}(t)
    =
    (\mu_x,\mu_y,\mu_z)^{\top}
    +(t-\mu_t)\frac{V}{W}.
\end{equation}
The marginal distribution $p(t)$ is also Gaussian:
$p(t)=\mathcal{N}(\mu_t,W)$.

\subsection{Motion-Expressive 4D Gaussian Representation}

\noindent\textbf{Extension of 4D Gaussians.}
The motion represented by conventional 4D Gaussians~\cite{4d-rotor2024Siggraph,Real-time2024ICLR} is limited because each Gaussian follows a linear trajectory
\begin{equation}
    \bm{\mu}(t)
    =
    (\mu_x,\mu_y,\mu_z)^{\top}
    +(t-\mu_t)\frac{V}{W}
\end{equation}
in space.
As a result, complex motion may be divided among multiple temporally localized Gaussians, leading to redundant modeling or degeneration toward per-frame reconstruction.

To address this issue, we extend the definition of $\bm{\mu}(t)$ in Eq.~(\ref{eq:projected 3D gaussian}) from a linear function to a combination of polynomials and Fourier series.
Let $\tau=t-\mu_t$ denote the time relative to the temporal center of each Gaussian.
The extended mean is defined as
\begin{equation}
\begin{aligned}
    \bm{\mu}(t)
    =
    (\mu_x,\mu_y,\mu_z)^{\top}
    &+
    \sum_{n=1}^{N}
    \bm{a}_{n}^{\mu}\tau^{n} \\
    &+
    \sum_{l=1}^{L}
    \left[
        \bm{b}_{l}^{\mu}\cos(l\tau)
        +
        \bm{c}_{l}^{\mu}\sin(l\tau)
    \right].
\end{aligned}
\label{eq:extended 4D means}
\end{equation}
Here,
$\bm{a}_{n}^{\mu},
\bm{b}_{l}^{\mu},
\bm{c}_{l}^{\mu}\in\mathbb{R}^{3}$
are learnable parameters.
The polynomial components capture smooth motion trends, while the Fourier components provide additional capacity for rapid motion variations.

Similarly, for
$\Sigma_{\text{3D}}=RSS^{\top}R^{\top}$,
we replace the constant rotation with a time-varying quaternion:
\begin{equation}
\begin{aligned}
    \bm{q}(t)
    =
    \bm{q}_{0}
    &+
    \sum_{n=0}^{N}
    \bm{a}_{n}^{q}\tau^{n} \\
    &+
    \sum_{l=1}^{L}
    \left[
        \bm{b}_{l}^{q}\cos(l\tau)
        +
        \bm{c}_{l}^{q}\sin(l\tau)
    \right],
\end{aligned}
\end{equation}
where
$\bm{a}_{n}^{q},
\bm{b}_{l}^{q},
\bm{c}_{l}^{q}\in\mathbb{R}^{4}$
are learnable parameters.
The scale matrix $S$ remains constant.
In this manner, each Gaussian can represent a nonlinear trajectory and a time-varying covariance orientation, resulting in a more motion-expressive 4D Gaussian representation.

To use this representation for motion supervision, we associate each Gaussian with the velocity of its center.
Differentiating Eq.~(\ref{eq:extended 4D means}) gives
\begin{equation}
\begin{aligned}
    \bm{v}(t)
    &=
    \frac{\partial\bm{\mu}(t)}{\partial t} \\
    &=
    \sum_{n=1}^{N}
    n\bm{a}_{n}^{\mu}\tau^{n-1} \\
    &\quad+
    \sum_{l=1}^{L}
    l\left[
        -\bm{b}_{l}^{\mu}\sin(l\tau)
        +
        \bm{c}_{l}^{\mu}\cos(l\tau)
    \right].
\end{aligned}
\label{eq:gaussian_velocity}
\end{equation}
These per-Gaussian velocities are subsequently projected and splatted to construct a dense motion field in image space.

\noindent\textbf{Flow Splatting.}
During rendering, the color of a pixel is computed by blending visible 3D Gaussians sorted according to their depth, as formulated in Eq.~(\ref{eq:color rendering}).
Similar strategies are employed for depth and feature rendering~\cite{Langsplat2024CVPR, Feature3dgs2024CVPR}.
Inspired by these methods, we directly splat the velocity of each Gaussian using the same rendering process:
\begin{equation}
    V(t)
    =
    \sum_{i\in N}
    \bm{v}_{i}(t)\alpha_{i}
    \prod_{j=1}^{i-1}(1-\alpha_{j}),
\end{equation}
where $\bm{v}_{i}(t)$ is the velocity of the $i$-th Gaussian at time $t$.

However, this rendering result cannot be directly supervised by optical flow because optical flow is also affected by camera motion.
Therefore, we take the camera parameters into account to eliminate projection differences caused by changes in camera pose.
Given camera parameters
$\bm{k}_t$ and $\bm{k}_{t+1}$
at times $t$ and $t+1$, respectively, the velocity on the 2D image plane is represented as
\begin{equation}
\begin{aligned}
    \hat{\bm{v}}(t)
    &=
    \frac{
        \psi_{\text{proj}}
        (\bm{\mu}(t),\bm{k}_{t+1})
        -
        \psi_{\text{proj}}
        (\bm{\mu}(t),\bm{k}_{t})
    }{\Delta t} \\
    &\quad+
    \psi_{\text{proj}}
    (\bm{v}(t),\bm{k}_{t+1}),
\end{aligned}
\end{equation}
where $\psi_{\text{proj}}$ is the projection operation and
$\hat{\bm{v}}(t)\in\mathbb{R}^{2}$.
We then splat the optical flow through differentiable rasterization:
\begin{equation}
    \hat{V}
    =
    \sum_{i\in N}
    \hat{\bm{v}}_{i}(t)\alpha_{i}
    \prod_{j=1}^{i-1}(1-\alpha_{j}).
    \label{eq:flow splatting}
\end{equation}
The rendered result corresponds to the optical flow of the input video and can be supervised using predictions from off-the-shelf models.

\subsection{Training}

We leverage a set of data-driven priors obtained from off-the-shelf models, including depth and camera estimation~\cite{Megasam2024arXiv}, mask prediction~\cite{Sam2023ICCV}, point tracking~\cite{Tapir2023ICCV}, and optical flow prediction~\cite{RAFT2020ECCV}.

\noindent\textbf{Initialization.}
Given a monocular video
$\{I_i\}_{i=1}^{T}$,
we first estimate the depth maps
$\{D_i\}_{i=1}^{T}$,
the corresponding camera parameters
$\{\bm{k}_i\}_{i=1}^{T}$,
and foreground masks
$\{M_i\}_{i=1}^{T}$.
We adopt different initialization strategies for the moving foreground and static background.

For the static background, we initialize
$\lambda=10^{-6}$
and
$\mu_t=T/2$
to maintain a similar temporal density across all frames.
The 3D locations are initialized by unprojecting pixels into 3D space using the aligned depth maps, while the learnable coefficients
$\bm{a}_{n}^{\mu}$,
$\bm{b}_{l}^{\mu}$,
$\bm{c}_{l}^{\mu}$,
$\bm{a}_{n}^{q}$,
$\bm{b}_{l}^{q}$, and
$\bm{c}_{l}^{q}$
are initialized to zero.

For the moving foreground, we combine the lifted 2D tracks with the aligned depth maps to obtain initial 3D trajectory observations.
We uniformly sample $\hat{T}$ frames
$\{I_{s_i}\}_{i=1}^{\hat{T}}$
from the video and initialize $\hat{T}$ Gaussians for each trajectory.

We initialize $\lambda$ following the Nyquist--Shannon sampling theorem.
Since the temporal sampling frequency is
$\nu_{\text{s}}=\hat{T}$,
the highest recoverable frequency is
\begin{equation}
    \nu_{\text{r}}
    =
    \frac{\nu_{\text{s}}}{2}
    =
    \frac{\hat{T}}{2}.
\end{equation}
The Fourier transform of $p(t)$ is formulated as
\begin{equation}
\begin{aligned}
    p(t)
    &=
    \frac{1}{\sqrt{2\pi}\sigma}
    e^{-\frac{t^2}{2\sigma^2}}, \\
    F(\nu)
    &=
    e^{-\frac{\nu^2}
    {2(1/(2\pi\sigma))^2}},
\end{aligned}
\end{equation}
where $\sigma=1/\sqrt{\lambda}$.
To retain at least $68.4\%$ of $F(\nu)$, we set
\begin{equation}
    \nu_{\text{r}}
    >
    \frac{1}{2\pi\sigma}.
\end{equation}
Thus, $\lambda$ is initialized as
$\hat{T}^{2}\pi^{2}$.
We further use neighboring locations along each trajectory to initialize the low-order coefficients of $\bm{\mu}(t)$.

\noindent\textbf{Optimization.}
We introduce two losses to supervise the learning of the color and velocity fields.
At timestamp $t_0$, we compute the conditional distribution
$p(\bm{x}\vert t_0)$,
which determines the location and shape of the 3D primitives, and the marginal distribution
$p(t_0)$,
which controls their temporal density.
Following the conventional splatting scheme in Eq.~(\ref{eq:color rendering}), we render all frames
$\{\tilde{I}_i\}_{i=1}^{T}$.
The color loss is formulated as
\begin{equation}
\begin{aligned}
    \mathcal{L}_{\text{color}}
    =
    \frac{1}{T}
    \sum_{i=1}^{T}
    \Big[
        &(1-\gamma_1)
        \mathcal{L}_{1}
        (I_i,\tilde{I}_i) \\
        &+
        \gamma_1
        \mathcal{L}_{\text{SSIM}}
        (I_i,\tilde{I}_i)
    \Big],
\end{aligned}
\end{equation}
where
$\mathcal{L}_{1}$
and
$\mathcal{L}_{\text{SSIM}}$
denote the $L_1$ and SSIM losses, respectively, and
$\gamma_1$ is a hyperparameter.

To supervise dynamic information, we first estimate optical flow
$\{V_i\}_{i=1}^{T-1}$
from the input video using an off-the-shelf model.
Based on our Flow Splatting strategy, we render optical flow
$\{\tilde{V}_i\}_{i=1}^{T-1}$
using Eq.~(\ref{eq:flow splatting}).
The rendering loss for the velocity field is
\begin{equation}
    \mathcal{L}_{\text{flow}}
    =
    \frac{1}{T-1}
    \sum_{i=1}^{T-1}
    \mathcal{L}_{1}
    (V_i,\tilde{V}_i).
\end{equation}

Because Gaussians are optimized independently, their motion may become inconsistent with that of their spatial neighbors, which does not align with physical scene motion~\cite{Gaussian-flow2024CVPR}.
We therefore introduce a velocity consistency regularizer based on local rigidity.
This regularizer encourages nearby Gaussians to exhibit similar motion trends during optimization.

Assume that the dynamic scene comprises $N$ primitives
$\mathcal{G}=\{G_i\}_{i=1}^{N}$.
At timestamp $t_0$, we determine the active Gaussian set
$\mathcal{G}_{\text{a}}\subset\mathcal{G}$
using the marginal condition
$p(t_0)>\tau$,
where $\tau$ is a threshold.
We then use the kNN algorithm to compute the $k$ nearest neighbors
$\mathcal{N}_i$
for each
$G_i\in\mathcal{G}_{\text{a}}$.
The rigid loss is defined as
\begin{equation}
\begin{aligned}
    \mathcal{L}_{\text{rigid}}
    &=
    \frac{1}{k|\mathcal{G}_{\text{a}}|}
    \sum_{G_i\in\mathcal{G}_{\text{a}}}
    \sum_{G_j\in\mathcal{N}_i} \\
    &\quad
    w_{i,j}
    \Vert\bm{v}_{i}-\bm{v}_{j}\Vert,
\end{aligned}
\end{equation}
where
\begin{equation}
    w_{i,j}
    =
    \exp\left(
    -\beta
    \Vert\bm{\mu}_{i}-\bm{\mu}_{j}\Vert
    \right)
\end{equation}
is a weighting factor based on the spatial distance between a Gaussian pair.

The total velocity loss is
\begin{equation}
    \mathcal{L}_{\text{velocity}}
    =
    \gamma_2\mathcal{L}_{\text{flow}}
    +
    \gamma_3\mathcal{L}_{\text{rigid}},
\end{equation}
where $\gamma_2$ and $\gamma_3$ are hyperparameters.
The final optimization objective is
\begin{equation}
    \mathcal{L}
    =
    \mathcal{L}_{\text{color}}
    +
    \mathcal{L}_{\text{velocity}}.
\end{equation}

\section{Experiments}\label{sec:experiments}

\begin{figure*}[t]
    \centering
    \includegraphics[width=\linewidth]{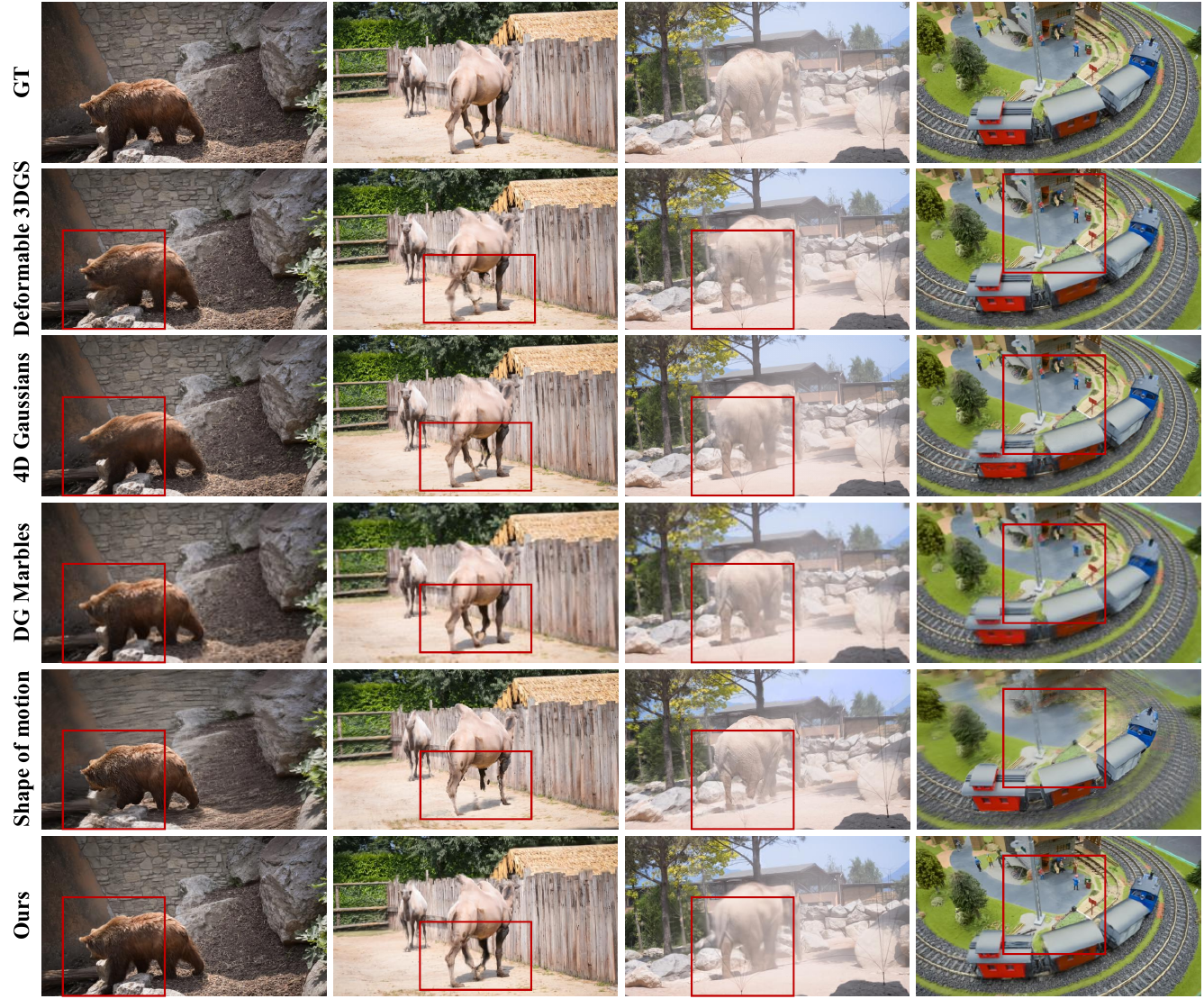}
    \caption{\textbf{Qualitative comparison on the DAVIS dataset.}
    We compare our method with previous methods for novel-view and novel-time synthesis.}
    \label{fig:main_results}
\end{figure*}

\subsection{Experimental Settings}

\noindent\textbf{Implementation Details.}
All experiments are conducted on an NVIDIA RTX A6000 GPU.
Although each 4D Gaussian theoretically extends infinitely in time, we filter out Gaussians with a temporal marginal $p(t)<0.05$ when rendering the view at time $t$.
We train our method using the Adam optimizer for a total of 1,500 iterations and densify the 4D Gaussians every 100 iterations.
For optimization, we set $\gamma_{1}=0.2$, $\gamma_{2}=0.03$, $\gamma_{3}=0.5$, and $\beta=100$, and choose $k=20$ for the local velocity consistency loss $\mathcal{L}_{\text{rigid}}$.
We uniformly sample 7 frames for foreground initialization.
The orders of the polynomial and Fourier components are set to 6 for $\bm{\mu}(t)$ and 3 for $\bm{q}(t)$.

\noindent\textbf{Datasets.}
The DAVIS dataset~\cite{Davis2017arXiv} contains real-world videos of 30 to 100 frames with various scenarios and motion dynamics.
We uniformly select one out of every ten frames as the test set, creating a challenging setting in which both the timestamps and viewpoints used for evaluation are novel.
We report per-scene reconstruction quality on Bear, Breakdance-flare, Camel, Train, and Elephant.
The NVIDIA Dynamic Scenes dataset~\cite{NVIDIA2020CVPR} consists of seven videos, including Balloon1, Balloon2, Jumping, Playground, Skating, Truck, and Umbrella.
Each scene contains sequences of 90 to 200 frames captured using a rig of 12 calibrated cameras.
Following the setting of Gaussian Marbles~\cite{Marbles2024Siggraph}, we use the video stream from camera 4 for training and the streams from cameras 3, 5, and 6 for evaluation.

\noindent\textbf{Baselines.}
We compare our method with both NeRF-based methods~\cite{T-Nerf2022NIPS,Hypernerf2021arXiv} and concurrent Gaussian-based methods.
T-NeRF~\cite{T-Nerf2022NIPS} represents a time-varying neural radiance field conditioned on time, while HyperNeRF~\cite{Hypernerf2021arXiv} models scene dynamics using a deformation field that maps observations to a canonical space.
Deformable 3DGS~\cite{Deformable3DGS2024CVPR} uses an MLP-based deformation field, while 4D Gaussians~\cite{4DGS2024CVPR} employs HexPlane features and a lightweight MLP to predict Gaussian deformations.
DG Marbles~\cite{Marbles2024Siggraph} uses Gaussian marbles and a hierarchical learning strategy, while Shape of Motion~\cite{Shapeofmotion2024arXiv} relies on an explicit motion representation.
We use 4DGS~\cite{Real-time2024ICLR}, which directly represents dynamic scenes using 4D Gaussian primitives, as our baseline.
Starting from this baseline, our method improves the motion capacity of each primitive using means and covariances that vary over time, and introduces Flow Splatting to provide dense motion supervision.

\subsection{Main Results}

\begin{table*}[t]
    \centering
    \caption{\textbf{Quantitative comparison on the NVIDIA dataset.}
    We report the number of Gaussians (k) for Gaussian-based methods.
    For 4DGS~\cite{Real-time2024ICLR} and our method, we report the
    number of active 4D Gaussians in each frame, with the total number
    of 4D Gaussians shown in parentheses.}

    \setlength{\tabcolsep}{1.2mm}
    \begin{tabular}{l|ccc|ccc}
        \toprule
        Methods
        & PSNR & SSIM & LPIPS
        & Training & FPS & Gaussians \\
        \midrule

        T-NeRF~\cite{T-Nerf2022NIPS}
        & 20.76 & 0.59 & 0.17
        & $>20$h & $<1$ & \cellcolor{gray!10}{} \\

        HyperNeRF~\cite{Hypernerf2021arXiv}
        & 20.05 & 0.57 & 0.18
        & $>20$h & $<1$ & \cellcolor{gray!10}{} \\

        \midrule

        Deformable 3DGS~\cite{Deformable3DGS2024CVPR}
        & 16.64 & 0.48 & 0.31
        & 2.2h & 12.4 & 694 \\

        4D Gaussians~\cite{4DGS2024CVPR}
        & 17.69 & 0.48 & 0.38
        & 1.7h & 80.1 & 204 \\

        DG Marbles~\cite{Marbles2024Siggraph}
        & 22.36 & 0.66 & 0.15
        & 3.1h & 175.0 & 120 \\

        Shape of Motion~\cite{Shapeofmotion2024arXiv}
        & 22.07 & 0.63 & 0.15
        & 2.3h & 40.3 & 648 \\


        \rowcolor{gray!20}
        4DGS~\cite{Real-time2024ICLR}
        & 16.91 & 0.37 & 0.35
        & 1.2h & 259.6 & 119 (1088) \\

        \rowcolor{gray!20}
        Ours
        & \makebox[24mm][c]{%
            \makebox[0pt][c]{22.72}%
            \makebox[0pt][l]{\hspace{5.3mm}%
            \textcolor{green}{\scriptsize$+5.81$}}}
        & \makebox[22mm][c]{%
            \makebox[0pt][c]{0.72}%
            \makebox[0pt][l]{\hspace{4.4mm}%
            \textcolor{green}{\scriptsize$+0.35$}}}
        & \makebox[22mm][c]{%
            \makebox[0pt][c]{0.16}%
            \makebox[0pt][l]{\hspace{4.4mm}%
            \textcolor{green}{\scriptsize$-0.19$}}}
        & 1.0h & 330.6 & 91 (196) \\

        \bottomrule
    \end{tabular}

    \label{tab:nvidia}
\end{table*}

    

\begin{table*}[t]
    \centering
    \caption{\textbf{Quantitative comparison on the DAVIS dataset.}
    We report PSNR for each test scene and the average PSNR, SSIM, and LPIPS over all scenes.}

    \setlength{\tabcolsep}{2mm}
    \begin{tabular}{l|ccccc|ccc}
        \toprule
        Methods
        & bear & flare & camel & elephant & train
        & PSNR & SSIM & LPIPS \\
        \midrule

        Deformable 3DGS~\cite{Deformable3DGS2024CVPR}
        & \cellcolor{yellow!20}{}28.74
        & 24.65
        & 25.79
        & \cellcolor{yellow!20}{}31.02
        & 21.52
        & 26.34
        & 0.81
        & \cellcolor{yellow!20}{}0.15 \\

        4D Gaussians~\cite{4DGS2024CVPR}
        & \cellcolor{orange!20}{}29.49
        & \cellcolor{yellow!20}{}24.68
        & \cellcolor{yellow!20}{}26.62
        & \cellcolor{orange!20}{}31.66
        & \cellcolor{yellow!20}{}23.60
        & \cellcolor{orange!20}{}27.21
        & \cellcolor{yellow!20}{}0.82
        & 0.22 \\

        DG Marbles~\cite{Marbles2024Siggraph}
        & 28.15
        & \cellcolor{orange!20}{}25.26
        & \cellcolor{orange!20}{}28.14
        & 28.48
        & \cellcolor{orange!20}{}23.90
        & \cellcolor{yellow!20}{}26.79
        & \cellcolor{orange!20}{}0.86
        & \cellcolor{red!20}{}0.07 \\

        Shape of Motion~\cite{Shapeofmotion2024arXiv}
        & 23.05
        & 23.90
        & 23.77
        & 26.81
        & 20.10
        & 23.53
        & 0.70
        & 0.21 \\


        Ours
        & \cellcolor{red!20}{}31.88
        & \cellcolor{red!20}{}28.48
        & \cellcolor{red!20}{}29.27
        & \cellcolor{red!20}{}33.45
        & \cellcolor{red!20}{}29.51
        & \cellcolor{red!20}{}30.52
        & \cellcolor{red!20}{}0.92
        & \cellcolor{orange!20}{}0.10 \\

        \bottomrule
    \end{tabular}

    \label{tab:davis}
\end{table*}

\noindent\textbf{Quantitative and Qualitative Results.}
We conduct novel view synthesis experiments on the NVIDIA dataset~\cite{NVIDIA2020CVPR} and report the quantitative results in Table~\ref{tab:nvidia}.
Compared with our baseline~\cite{Real-time2024ICLR}, our method improves PSNR by 5.81 and SSIM by 0.35, while reducing LPIPS by 0.19.
It also requires less training time and achieves a higher rendering speed.
Compared with DG Marbles~\cite{Marbles2024Siggraph}, the strongest competing method in terms of PSNR and SSIM, our method improves PSNR by 0.36 and SSIM by 0.06, while reducing the training time from 3.1 hours to 1.0 hour.

We further evaluate our method on DAVIS.
As shown in Table~\ref{tab:davis}, our method achieves the highest PSNR on every test scene, as well as the best average PSNR and SSIM.
More specifically, it improves over the deformation-based 4D Gaussians method~\cite{4DGS2024CVPR} by 3.31 in PSNR and 0.10 in SSIM, while reducing LPIPS by 0.12.
It also outperforms the trajectory-based DG Marbles method~\cite{Marbles2024Siggraph} by 3.73 in PSNR and 0.06 in SSIM, while DG Marbles retains a lower LPIPS.
Qualitative results are shown in Figure~\ref{fig:main_results}.
Our method preserves more scene details, while competing methods may exhibit missing local details, floaters, or incorrect dynamics.

\noindent\textbf{Efficiency Analysis.}
Our method provides a favorable efficiency advantage over both NeRF-based and Gaussian-based methods.
Although NeRF-based methods~\cite{T-Nerf2022NIPS,Hypernerf2021arXiv} can produce high-quality images, their implicit representations require long optimization times and render at less than one frame per second.
Deformation-based Gaussian methods~\cite{Deformable3DGS2024CVPR,4DGS2024CVPR} require additional network inference to predict Gaussian deformations at novel timestamps.
Our method renders at 330.6 FPS, compared with 12.4 FPS for Deformable 3DGS and 80.1 FPS for 4D Gaussians.
The compared trajectory-based methods require between 2.3 and 3.1 hours of training, whereas our method converges in 1.0 hour.

The compactness of our representation further supports the benefit of improving the motion capacity of individual 4D Gaussian primitives.
Our method uses 91 active and 196 total Gaussians, compared with 119 active and 1,088 total Gaussians used by the conventional 4DGS baseline~\cite{Real-time2024ICLR}.
This substantial reduction indicates that complex motion can be represented using fewer primitives, reducing redundant scene modeling.

\begin{table*}[!t]
    \centering
    \caption{\textbf{Ablation study of the proposed representation and training strategies.}
    $\bm{\mu}(t)$ and $\bm{q}(t)$ denote the proposed temporal extensions of the 4D Gaussian representation.
    Init. denotes the proposed initialization strategy.
    For the representation ablation, we retain all training strategies;
    for the training ablation, we use the complete motion-expressive 4D Gaussian representation.}
    \label{tab:ablation}

    \begin{subtable}[t]{0.48\textwidth}
        \centering
        \caption{4D Gaussian representation}
        \label{tab:ablation-representation}

        \begin{tabular*}{\linewidth}
        {@{\extracolsep{\fill}}cc|ccc@{}}
            \toprule
            $\bm{\mu}(t)$
            & $\bm{q}(t)$
            & PSNR & SSIM & LPIPS \\
            \midrule

            \XSolidBrush
            & \XSolidBrush
            & 27.17 & 0.893 & 0.091 \\

            \CheckmarkBold
            & \XSolidBrush
            & 27.31 & 0.893 & 0.090 \\

            \XSolidBrush
            & \CheckmarkBold
            & 27.24 & 0.893 & 0.092 \\

            \CheckmarkBold
            & \CheckmarkBold
            & 28.02 & 0.910 & 0.082 \\

            \bottomrule
        \end{tabular*}
    \end{subtable}
    \hfill
    \begin{subtable}[t]{0.48\textwidth}
        \centering
        \caption{Training strategy}
        \label{tab:ablation-training}

        \begin{tabular*}{\linewidth}
        {@{\extracolsep{\fill}}ccc|ccc@{}}
            \toprule
            Init.
            & $\mathcal{L}_{\text{flow}}$
            & $\mathcal{L}_{\text{rigid}}$
            & PSNR & SSIM & LPIPS \\
            \midrule

            \XSolidBrush
            & \CheckmarkBold
            & \CheckmarkBold
            & 25.97 & 0.829 & 0.150 \\

            \CheckmarkBold
            & \XSolidBrush
            & \XSolidBrush
            & 24.89 & 0.776 & 0.191 \\

            \CheckmarkBold
            & \CheckmarkBold
            & \XSolidBrush
            & 26.87 & 0.861 & 0.115 \\

            \CheckmarkBold
            & \CheckmarkBold
            & \CheckmarkBold
            & 28.02 & 0.910 & 0.082 \\

            \bottomrule
        \end{tabular*}
    \end{subtable}
\end{table*}

\begin{figure*}[!t]
    \centering
    \includegraphics[width=\textwidth]{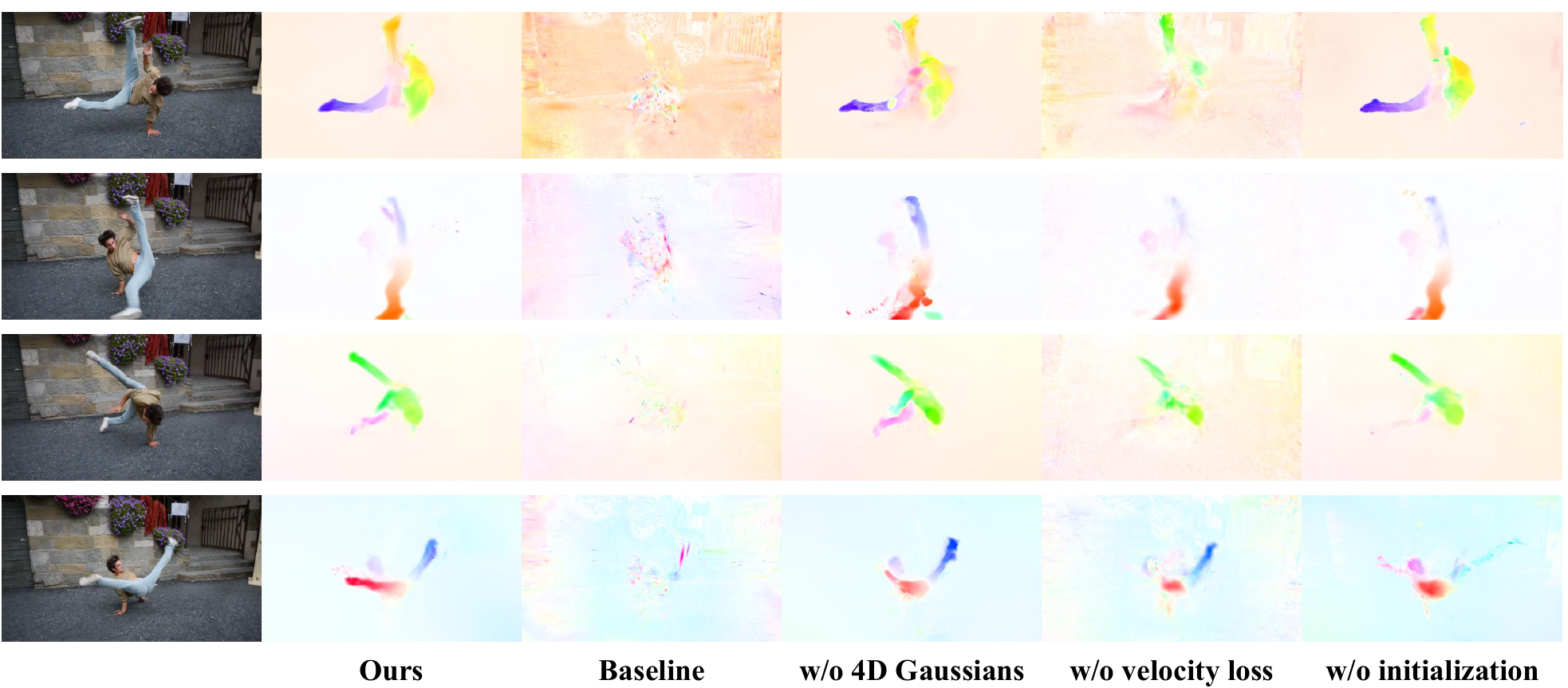}
    \caption{\textbf{Qualitative comparison of rendered optical flow.}
    All variants are rendered using Flow Splatting.
    ``Baseline'' denotes conventional 4DGS~\cite{Real-time2024ICLR};
    ``w/o 4D Gaussian'' removes our polynomial and Fourier extensions;
    ``w/o velocity loss'' removes velocity-field supervision;
    and ``w/o initialization'' uses the baseline random initialization.}
    \label{fig:optical_flow}
    \vspace{-2mm}
\end{figure*}

\subsection{Ablation Study and Analysis}

We conduct ablation studies to evaluate the proposed motion-expressive 4D Gaussian representation and the corresponding training strategies.

\noindent\textbf{4D Gaussian Representation.}
We first consider the conventional representation used in our baseline~\cite{Real-time2024ICLR}, without the polynomial and Fourier components.
We then extend either $\bm{\mu}(t)$ or $\bm{q}(t)$ with the proposed time-varying components and finally evaluate the complete representation.
As shown in Table~\ref{tab:ablation-representation}, removing the temporal extension of $\bm{\mu}(t)$ or $\bm{q}(t)$ decreases PSNR by 0.78 or 0.71, respectively.
Removing both extensions further reduces PSNR from 28.02 to 27.17.
These results demonstrate that both the nonlinear trajectory and the covariance orientation that varies over time contribute to the motion capacity of the proposed representation.

We visualize the rendered optical flow in Figure~\ref{fig:optical_flow}.
The conventional 4DGS baseline fails to recover accurate motion in both static and dynamic regions.
In comparison, the proposed representation reconstructs foreground motion more completely using its extended temporal parameterization.

\noindent\textbf{Training Strategy.}
We retain the color rendering loss and ablate the initialization strategy and the two velocity loss terms, $\mathcal{L}_{\text{flow}}$ and $\mathcal{L}_{\text{rigid}}$.
As shown in Table~\ref{tab:ablation-training}, removing the proposed initialization decreases PSNR by 2.05.
Starting from the model trained without either velocity loss, adding Flow Splatting supervision improves PSNR from 24.89 to 26.87, corresponding to a gain of 1.98.
Adding the local velocity consistency regularizer further improves PSNR from 26.87 to 28.02, providing an additional gain of 1.15.
Together, the two velocity loss terms improve PSNR by 3.13.
The visualizations in Figure~\ref{fig:optical_flow} further demonstrate the importance of these training strategies.

\section{Conclusion}\label{sec:conclusion}


In this paper, we propose a motion-expressive 4D Gaussian representation for dynamic scene reconstruction from monocular videos. Its time-varying means and covariances enable each primitive to model complex nonlinear motion. We analytically derive Gaussian velocities and introduce Flow Splatting to render optical flow while accounting for camera motion, enabling dense motion supervision. This formulation bridges 3D Gaussian dynamics and observable 2D motion cues, effectively reducing the ambiguity inherent in monocular dynamic scene reconstruction. We further use the proposed initialization and a velocity consistency regularizer to stabilize optimization. Experiments demonstrate favorable reconstruction quality and efficiency.

\noindent\textbf{Limitations and future works.}
Although our method produces compelling results, it still has limitations.
Compared with feed-forward methods, our optimization-based approach requires per-scene training.
The reconstruction pipeline does not use generative priors and therefore cannot recover unseen regions.
Its motion supervision also depends on estimated optical flow, which may be unreliable under severe occlusions or large displacements.
Furthermore, our method focuses on appearance and motion without explicitly constraining scene geometry.
Future work will explore feed-forward reconstruction, generative priors, and geometry-aware modeling.


{
\small
\bibliographystyle{ieeenat_fullname}
\bibliography{main}

@article{3DGS2023ToG,
  title={3d gaussian splatting for real-time radiance field rendering.},
  author={Kerbl, Bernhard and Kopanas, Georgios and Leimk{\"u}hler, Thomas and Drettakis, George},
  journal={ACM Trans. Graph.},
  volume={42},
  number={4},
  pages={139--1},
  year={2023}
}

@article{Real-time2024ICLR,
  title={Real-time photorealistic dynamic scene representation and rendering with 4d gaussian splatting},
  author={Yang, Zeyu and Yang, Hongye and Pan, Zijie and Zhang, Li},
  journal={ICLR},
  year={2024}
}

@inproceedings{4d-rotor2024Siggraph,
  title={4d-rotor gaussian splatting: towards efficient novel view synthesis for dynamic scenes},
  author={Duan, Yuanxing and Wei, Fangyin and Dai, Qiyu and He, Yuhang and Chen, Wenzheng and Chen, Baoquan},
  booktitle={ACM SIGGRAPH 2024 Conference Papers},
  pages={1--11},
  year={2024}
}

@inproceedings{Vr-gs2024Siggraph,
  title={Vr-gs: A physical dynamics-aware interactive gaussian splatting system in virtual reality},
  author={Jiang, Ying and Yu, Chang and Xie, Tianyi and Li, Xuan and Feng, Yutao and Wang, Huamin and Li, Minchen and Lau, Henry and Gao, Feng and Yang, Yin and others},
  booktitle={ACM SIGGRAPH 2024 Conference Papers},
  pages={1--1},
  year={2024}
}

@article{Robot2022RAL,
  title={Vision-only robot navigation in a neural radiance world},
  author={Adamkiewicz, Michal and Chen, Timothy and Caccavale, Adam and Gardner, Rachel and Culbertson, Preston and Bohg, Jeannette and Schwager, Mac},
  journal={IEEE Robotics and Automation Letters},
  volume={7},
  number={2},
  pages={4606--4613},
  year={2022},
  publisher={IEEE}
}

@inproceedings{Drivinggaussian2024CVPR,
  title={Drivinggaussian: Composite gaussian splatting for surrounding dynamic autonomous driving scenes},
  author={Zhou, Xiaoyu and Lin, Zhiwei and Shan, Xiaojun and Wang, Yongtao and Sun, Deqing and Yang, Ming-Hsuan},
  booktitle={Proceedings of the IEEE/CVF conference on computer vision and pattern recognition},
  pages={21634--21643},
  year={2024}
}

@inproceedings{LFN2021NIPS,
  title={Light Field Networks: Neural Scene Representations with Single-Evaluation Rendering},
  author={Vincent Sitzmann and Semon Rezchikov and William T. Freeman and Joshua B. Tenenbaum and Fr{\'e}do Durand},
  booktitle={NeurIPS},
  year={2021},
  url={https://api.semanticscholar.org/CorpusID:235352518}
}

@article{NeRF2020ECCV,
  title={Nerf: Representing scenes as neural radiance fields for view synthesis},
  author={Mildenhall, Ben and Srinivasan, Pratul P and Tancik, Matthew and Barron, Jonathan T and Ramamoorthi, Ravi and Ng, Ren},
  journal={ECCV},
  year={2020},
}

@article{SRN2019NIPS,
  title={Scene representation networks: Continuous 3d-structure-aware neural scene representations},
  author={Sitzmann, Vincent and Zollhofer, Michael and Wetzstein, Gordon},
  journal={NeurIPS},
  volume={32},
  year={2019}
}

@inproceedings{Plenoxels2022CVPR,
  title={Plenoxels: Radiance fields without neural networks},
  author={Fridovich-Keil, Sara and Yu, Alex and Tancik, Matthew and Chen, Qinhong and Recht, Benjamin and Kanazawa, Angjoo},
  booktitle={CVPR},
  pages={5501--5510},
  year={2022}
}

@inproceedings{PlenOctree2021CVPR,
  title={Plenoctrees for real-time rendering of neural radiance fields},
  author={Yu, Alex and Li, Ruilong and Tancik, Matthew and Li, Hao and Ng, Ren and Kanazawa, Angjoo},
  booktitle={ICCV},
  pages={5752--5761},
  year={2021}
}

@inproceedings{EfficientNeRF2022CVPR,
  title={Efficientnerf efficient neural radiance fields},
  author={Hu, Tao and Liu, Shu and Chen, Yilun and Shen, Tiancheng and Jia, Jiaya},
  booktitle={CVPR},
  pages={12902--12911},
  year={2022}
}

@article{NSVF2020NIPS,
  title={Neural sparse voxel fields},
  author={Liu, Lingjie and Gu, Jiatao and Zaw Lin, Kyaw and Chua, Tat-Seng and Theobalt, Christian},
  journal={NeurIPS},
  volume={33},
  pages={15651--15663},
  year={2020}
}

@article{InstantNGP2022TOG,
  title={Instant neural graphics primitives with a multiresolution hash encoding},
  author={M{\"u}ller, Thomas and Evans, Alex and Schied, Christoph and Keller, Alexander},
  journal={ToG},
  volume={41},
  number={4},
  pages={1--15},
  year={2022},
  publisher={ACM New York, NY, USA}
}

@inproceedings{KiloNeRF2021ICCV,
  title={Kilonerf: Speeding up neural radiance fields with thousands of tiny mlps},
  author={Reiser, Christian and Peng, Songyou and Liao, Yiyi and Geiger, Andreas},
  booktitle={ICCV},
  pages={14335--14345},
  year={2021}
}

@inproceedings{FastNeRF2021ICCV,
  title={Fastnerf: High-fidelity neural rendering at 200fps},
  author={Garbin, Stephan J and Kowalski, Marek and Johnson, Matthew and Shotton, Jamie and Valentin, Julien},
  booktitle={ICCV},
  pages={14346--14355},
  year={2021}
}

@inproceedings{MipNeRF2021ICCV,
  title={Mip-nerf: A multiscale representation for anti-aliasing neural radiance fields},
  author={Barron, Jonathan T and Mildenhall, Ben and Tancik, Matthew and Hedman, Peter and Martin-Brualla, Ricardo and Srinivasan, Pratul P},
  booktitle={ICCV},
  pages={5855--5864},
  year={2021}
}

@inproceedings{Deformable3DGS2024CVPR,
  title={Deformable 3d gaussians for high-fidelity monocular dynamic scene reconstruction},
  author={Yang, Ziyi and Gao, Xinyu and Zhou, Wen and Jiao, Shaohui and Zhang, Yuqing and Jin, Xiaogang},
  booktitle={Proceedings of the IEEE/CVF conference on computer vision and pattern recognition},
  pages={20331--20341},
  year={2024}
}

@inproceedings{4DGS2024CVPR,
  title={4d gaussian splatting for real-time dynamic scene rendering},
  author={Wu, Guanjun and Yi, Taoran and Fang, Jiemin and Xie, Lingxi and Zhang, Xiaopeng and Wei, Wei and Liu, Wenyu and Tian, Qi and Wang, Xinggang},
  booktitle={Proceedings of the IEEE/CVF conference on computer vision and pattern recognition},
  pages={20310--20320},
  year={2024}
}

@inproceedings{Gaufre2025WACV,
  title={Gaufre: Gaussian deformation fields for real-time dynamic novel view synthesis},
  author={Liang, Yiqing and Khan, Numair and Li, Zhengqin and Nguyen-Phuoc, Thu and Lanman, Douglas and Tompkin, James and Xiao, Lei},
  booktitle={2025 IEEE/CVF Winter Conference on Applications of Computer Vision (WACV)},
  pages={2642--2652},
  year={2025},
  organization={IEEE}
}

@inproceedings{Dynamic3DGS20243DV,
  title={Dynamic 3d gaussians: Tracking by persistent dynamic view synthesis},
  author={Luiten, Jonathon and Kopanas, Georgios and Leibe, Bastian and Ramanan, Deva},
  booktitle={2024 International Conference on 3D Vision (3DV)},
  pages={800--809},
  year={2024},
  organization={IEEE}
}

@article{Shapeofmotion2024arXiv,
  title={Shape of motion: 4d reconstruction from a single video},
  author={Wang, Qianqian and Ye, Vickie and Gao, Hang and Austin, Jake and Li, Zhengqi and Kanazawa, Angjoo},
  journal={arXiv preprint arXiv:2407.13764},
  year={2024}
}

@article{Mosca2024arXiv,
  title={Mosca: Dynamic gaussian fusion from casual videos via 4d motion scaffolds},
  author={Lei, Jiahui and Weng, Yijia and Harley, Adam and Guibas, Leonidas and Daniilidis, Kostas},
  journal={arXiv preprint arXiv:2405.17421},
  year={2024}
}

@inproceedings{MoDGS2025ICLR,
  title={MoDGS: Dynamic Gaussian Splatting from Casually-captured Monocular Videos with Depth Priors},
  author={Qingming, LIU and Liu, Yuan and Wang, Jiepeng and Lyu, Xianqiang and Wang, Peng and Wang, Wenping and Hou, Junhui},
  booktitle={The Thirteenth International Conference on Learning Representations},
  year={2025}
}

@article{Motion-aware2024TCSVT,
  title={Motion-aware 3d gaussian splatting for efficient dynamic scene reconstruction},
  author={Guo, Zhiyang and Zhou, Wengang and Li, Li and Wang, Min and Li, Houqiang},
  journal={IEEE Transactions on Circuits and Systems for Video Technology},
  year={2024},
  publisher={IEEE}
}

@inproceedings{OccNet2019CVPR,
  title={Occupancy networks: Learning 3d reconstruction in function space},
  author={Mescheder, Lars and Oechsle, Michael and Niemeyer, Michael and Nowozin, Sebastian and Geiger, Andreas},
  booktitle={Proceedings of the IEEE/CVF conference on computer vision and pattern recognition},
  pages={4460--4470},
  year={2019}
}

@inproceedings{Deepsdf2019CVPR,
  title={Deepsdf: Learning continuous signed distance functions for shape representation},
  author={Park, Jeong Joon and Florence, Peter and Straub, Julian and Newcombe, Richard and Lovegrove, Steven},
  booktitle={Proceedings of the IEEE/CVF conference on computer vision and pattern recognition},
  pages={165--174},
  year={2019}
}

@article{MVNS2020NIPS,
  title={Multiview neural surface reconstruction by disentangling geometry and appearance},
  author={Yariv, Lior and Kasten, Yoni and Moran, Dror and Galun, Meirav and Atzmon, Matan and Ronen, Basri and Lipman, Yaron},
  journal={Advances in Neural Information Processing Systems},
  volume={33},
  pages={2492--2502},
  year={2020}
}

@article{pixelSplat2023arXiv,
  title={pixelsplat: 3d gaussian splats from image pairs for scalable generalizable 3d reconstruction},
  author={Charatan, David and Li, Sizhe and Tagliasacchi, Andrea and Sitzmann, Vincent},
  journal={arXiv preprint arXiv:2312.12337},
  year={2023}
}

@article{MVSplat2024arXiv,
  title={MVSplat: Efficient 3D Gaussian Splatting from Sparse Multi-View Images},
  author={Chen, Yuedong and Xu, Haofei and Zheng, Chuanxia and Zhuang, Bohan and Pollefeys, Marc and Geiger, Andreas and Cham, Tat-Jen and Cai, Jianfei},
  journal={arXiv preprint arXiv:2403.14627},
  year={2024}
}

@article{SplatterImage2023arXiv,
  title={Splatter image: Ultra-fast single-view 3d reconstruction},
  author={Szymanowicz, Stanislaw and Rupprecht, Christian and Vedaldi, Andrea},
  journal={arXiv preprint arXiv:2312.13150},
  year={2023}
}

@inproceedings{BlockNeRF2022CVPR,
  title={Block-nerf: Scalable large scene neural view synthesis},
  author={Tancik, Matthew and Casser, Vincent and Yan, Xinchen and Pradhan, Sabeek and Mildenhall, Ben and Srinivasan, Pratul P and Barron, Jonathan T and Kretzschmar, Henrik},
  booktitle={CVPR},
  pages={8248--8258},
  year={2022}
}

@inproceedings{MegaNeRF2022CVPR,
  title={Mega-nerf: Scalable construction of large-scale nerfs for virtual fly-throughs},
  author={Turki, Haithem and Ramanan, Deva and Satyanarayanan, Mahadev},
  booktitle={CVPR},
  pages={12922--12931},
  year={2022}
}

@inproceedings{GridNeRF2023CVPR,
  title={Grid-guided neural radiance fields for large urban scenes},
  author={Xu, Linning and Xiangli, Yuanbo and Peng, Sida and Pan, Xingang and Zhao, Nanxuan and Theobalt, Christian and Dai, Bo and Lin, Dahua},
  booktitle={CVPR},
  pages={8296--8306},
  year={2023}
}

@inproceedings{BungeeNeRF2022ECCV,
  title={Bungeenerf: Progressive neural radiance field for extreme multi-scale scene rendering},
  author={Xiangli, Yuanbo and Xu, Linning and Pan, Xingang and Zhao, Nanxuan and Rao, Anyi and Theobalt, Christian and Dai, Bo and Lin, Dahua},
  booktitle={ECCV},
  pages={106--122},
  year={2022},
  organization={Springer}
}

@inproceedings{D-NeRF2021CVPR,
  title={D-nerf: Neural radiance fields for dynamic scenes},
  author={Pumarola, Albert and Corona, Enric and Pons-Moll, Gerard and Moreno-Noguer, Francesc},
  booktitle={CVPR},
  pages={10318--10327},
  year={2021}
}

@inproceedings{DynNeRF2021ICCV,
    author    = {Gao, Chen and Saraf, Ayush and Kopf, Johannes and Huang, Jia-Bin},
    title     = {Dynamic View Synthesis from Dynamic Monocular Video},
    booktitle = {ICCV},
    year      = {2021}
}

@inproceedings{NRFlow2021CS,
  title={Neural radiance flow for 4d view synthesis and video processing},
  author={Du, Yilun and Zhang, Yinan and Yu, Hong-Xing and Tenenbaum, Joshua B and Wu, Jiajun},
  booktitle={ICCV},
  pages={14304--14314},
  year={2021},
  organization={IEEE Computer Society}
}

@inproceedings{niemeyer2022regnerf,
  title={Regnerf: Regularizing neural radiance fields for view synthesis from sparse inputs},
  author={Niemeyer, Michael and Barron, Jonathan T and Mildenhall, Ben and Sajjadi, Mehdi SM and Geiger, Andreas and Radwan, Noha},
  booktitle={CVPR},
  pages={5480--5490},
  year={2022}
}

@inproceedings{truong2023sparf,
  title={Sparf: Neural radiance fields from sparse and noisy poses. IEEE},
  author={Truong, Prune and Rakotosaona, Marie-Julie and Manhardt, Fabian and Tombari, Federico},
  booktitle={CVPR},
  volume={1},
  year={2023}
}

@inproceedings{wynn2023diffusionerf,
  title={Diffusionerf: Regularizing neural radiance fields with denoising diffusion models},
  author={Wynn, Jamie and Turmukhambetov, Daniyar},
  booktitle={CVPR},
  pages={4180--4189},
  year={2023}
}

@article{xu2023murf,
  title={MuRF: Multi-Baseline Radiance Fields},
  author={Xu, Haofei and Chen, Anpei and Chen, Yuedong and Sakaridis, Christos and Zhang, Yulun and Pollefeys, Marc and Geiger, Andreas and Yu, Fisher},
  journal={arXiv preprint arXiv:2312.04565},
  year={2023}
}

@article{lu2023scaffold,
  title={Scaffold-gs: Structured 3d gaussians for view-adaptive rendering},
  author={Lu, Tao and Yu, Mulin and Xu, Linning and Xiangli, Yuanbo and Wang, Limin and Lin, Dahua and Dai, Bo},
  journal={arXiv preprint arXiv:2312.00109},
  year={2023}
}

@article{navaneet2023compact3d,
  title={Compact3d: Compressing gaussian splat radiance field models with vector quantization},
  author={Navaneet, KL and Meibodi, Kossar Pourahmadi and Koohpayegani, Soroush Abbasi and Pirsiavash, Hamed},
  journal={arXiv preprint arXiv:2311.18159},
  year={2023}
}

@article{girish2023eagles,
  title={Eagles: Efficient accelerated 3d gaussians with lightweight encodings},
  author={Girish, Sharath and Gupta, Kamal and Shrivastava, Abhinav},
  journal={arXiv preprint arXiv:2312.04564},
  year={2023}
}

@article{fan2023lightgaussian,
  title={Lightgaussian: Unbounded 3d gaussian compression with 15x reduction and 200+ fps},
  author={Fan, Zhiwen and Wang, Kevin and Wen, Kairun and Zhu, Zehao and Xu, Dejia and Wang, Zhangyang},
  journal={arXiv preprint arXiv:2311.17245},
  year={2023}
}

@article{katsumata2023efficient,
  title={An efficient 3d gaussian representation for monocular/multi-view dynamic scenes},
  author={Katsumata, Kai and Vo, Duc Minh and Nakayama, Hideki},
  journal={arXiv preprint arXiv:2311.12897},
  year={2023}
}

@article{yan2023multi,
  title={Multi-scale 3d gaussian splatting for anti-aliased rendering},
  author={Yan, Zhiwen and Low, Weng Fei and Chen, Yu and Lee, Gim Hee},
  journal={arXiv preprint arXiv:2311.17089},
  year={2023}
}

@article{gao2023relightable,
  title={Relightable 3d gaussian: Real-time point cloud relighting with brdf decomposition and ray tracing},
  author={Gao, Jian and Gu, Chun and Lin, Youtian and Zhu, Hao and Cao, Xun and Zhang, Li and Yao, Yao},
  journal={arXiv preprint arXiv:2311.16043},
  year={2023}
}

@article{jiang2023gaussianshader,
  title={GaussianShader: 3D Gaussian Splatting with Shading Functions for Reflective Surfaces},
  author={Jiang, Yingwenqi and Tu, Jiadong and Liu, Yuan and Gao, Xifeng and Long, Xiaoxiao and Wang, Wenping and Ma, Yuexin},
  journal={arXiv preprint arXiv:2311.17977},
  year={2023}
}

@article{liang2023gs,
  title={Gs-ir: 3d gaussian splatting for inverse rendering},
  author={Liang, Zhihao and Zhang, Qi and Feng, Ying and Shan, Ying and Jia, Kui},
  journal={arXiv preprint arXiv:2311.16473},
  year={2023}
}

@article{GGN2024NIPS,
  title={Gaussian graph network: Learning efficient and generalizable gaussian representations from multi-view images},
  author={Zhang, Shengjun and Fei, Xin and Liu, Fangfu and Song, Haixu and Duan, Yueqi},
  journal={Advances in Neural Information Processing Systems},
  volume={37},
  pages={50361--50380},
  year={2024}
}

@inproceedings{Neural3d2022CVPR,
  title={Neural 3d video synthesis from multi-view video},
  author={Li, Tianye and Slavcheva, Mira and Zollhoefer, Michael and Green, Simon and Lassner, Christoph and Kim, Changil and Schmidt, Tanner and Lovegrove, Steven and Goesele, Michael and Newcombe, Richard and others},
  booktitle={Proceedings of the IEEE/CVF conference on computer vision and pattern recognition},
  pages={5521--5531},
  year={2022}
}

@inproceedings{NeuralSceneFlow2021CVPR,
  title={Neural scene flow fields for space-time view synthesis of dynamic scenes},
  author={Li, Zhengqi and Niklaus, Simon and Snavely, Noah and Wang, Oliver},
  booktitle={Proceedings of the IEEE/CVF Conference on Computer Vision and Pattern Recognition},
  pages={6498--6508},
  year={2021}
}

@inproceedings{Nerfies2021ICCV,
  title={Nerfies: Deformable neural radiance fields},
  author={Park, Keunhong and Sinha, Utkarsh and Barron, Jonathan T and Bouaziz, Sofien and Goldman, Dan B and Seitz, Steven M and Martin-Brualla, Ricardo},
  booktitle={Proceedings of the IEEE/CVF international conference on computer vision},
  pages={5865--5874},
  year={2021}
}

@article{Hypernerf2021arXiv,
  title={Hypernerf: A higher-dimensional representation for topologically varying neural radiance fields},
  author={Park, Keunhong and Sinha, Utkarsh and Hedman, Peter and Barron, Jonathan T and Bouaziz, Sofien and Goldman, Dan B and Martin-Brualla, Ricardo and Seitz, Steven M},
  journal={arXiv preprint arXiv:2106.13228},
  year={2021}
}

@inproceedings{Space-time2021CVPR,
  title={Space-time neural irradiance fields for free-viewpoint video},
  author={Xian, Wenqi and Huang, Jia-Bin and Kopf, Johannes and Kim, Changil},
  booktitle={Proceedings of the IEEE/CVF conference on computer vision and pattern recognition},
  pages={9421--9431},
  year={2021}
}

@article{T-Nerf2022NIPS,
  title={Monocular dynamic view synthesis: A reality check},
  author={Gao, Hang and Li, Ruilong and Tulsiani, Shubham and Russell, Bryan and Kanazawa, Angjoo},
  journal={Advances in Neural Information Processing Systems},
  volume={35},
  pages={33768--33780},
  year={2022}
}

@inproceedings{Marbles2024Siggraph,
  title={Dynamic gaussian marbles for novel view synthesis of casual monocular videos},
  author={Stearns, Colton and Harley, Adam and Uy, Mikaela and Dubost, Florian and Tombari, Federico and Wetzstein, Gordon and Guibas, Leonidas},
  booktitle={SIGGRAPH Asia 2024 Conference Papers},
  pages={1--11},
  year={2024}
}

@inproceedings{NVIDIA2020CVPR,
  title={Novel view synthesis of dynamic scenes with globally coherent depths from a monocular camera},
  author={Yoon, Jae Shin and Kim, Kihwan and Gallo, Orazio and Park, Hyun Soo and Kautz, Jan},
  booktitle={Proceedings of the IEEE/CVF Conference on Computer Vision and Pattern Recognition},
  pages={5336--5345},
  year={2020}
}

@article{Davis2017arXiv,
  title={The 2017 davis challenge on video object segmentation},
  author={Pont-Tuset, Jordi and Perazzi, Federico and Caelles, Sergi and Arbel{\'a}ez, Pablo and Sorkine-Hornung, Alex and Van Gool, Luc},
  journal={arXiv preprint arXiv:1704.00675},
  year={2017}
}

@article{NeuralVolumes2019arXiv,
  title={Neural volumes: Learning dynamic renderable volumes from images},
  author={Lombardi, Stephen and Simon, Tomas and Saragih, Jason and Schwartz, Gabriel and Lehrmann, Andreas and Sheikh, Yaser},
  journal={arXiv preprint arXiv:1906.07751},
  year={2019}
}

@inproceedings{Hexplane2023CVPR,
  title={Hexplane: A fast representation for dynamic scenes},
  author={Cao, Ang and Johnson, Justin},
  booktitle={Proceedings of the IEEE/CVF Conference on Computer Vision and Pattern Recognition},
  pages={130--141},
  year={2023}
}

@inproceedings{Langsplat2024CVPR,
  title={Langsplat: 3d language gaussian splatting},
  author={Qin, Minghan and Li, Wanhua and Zhou, Jiawei and Wang, Haoqian and Pfister, Hanspeter},
  booktitle={Proceedings of the IEEE/CVF Conference on Computer Vision and Pattern Recognition},
  pages={20051--20060},
  year={2024}
}

@inproceedings{Feature3dgs2024CVPR,
  title={Feature 3dgs: Supercharging 3d gaussian splatting to enable distilled feature fields},
  author={Zhou, Shijie and Chang, Haoran and Jiang, Sicheng and Fan, Zhiwen and Zhu, Zehao and Xu, Dejia and Chari, Pradyumna and You, Suya and Wang, Zhangyang and Kadambi, Achuta},
  booktitle={Proceedings of the IEEE/CVF Conference on Computer Vision and Pattern Recognition},
  pages={21676--21685},
  year={2024}
}

@inproceedings{Tapir2023ICCV,
  title={Tapir: Tracking any point with per-frame initialization and temporal refinement},
  author={Doersch, Carl and Yang, Yi and Vecerik, Mel and Gokay, Dilara and Gupta, Ankush and Aytar, Yusuf and Carreira, Joao and Zisserman, Andrew},
  booktitle={Proceedings of the IEEE/CVF International Conference on Computer Vision},
  pages={10061--10072},
  year={2023}
}

@article{Megasam2024arXiv,
  title={Megasam: Accurate, fast, and robust structure and motion from casual dynamic videos},
  author={Li, Zhengqi and Tucker, Richard and Cole, Forrester and Wang, Qianqian and Jin, Linyi and Ye, Vickie and Kanazawa, Angjoo and Holynski, Aleksander and Snavely, Noah},
  journal={arXiv preprint arXiv:2412.04463},
  year={2024}
}

@inproceedings{Sam2023ICCV,
  title={Segment anything},
  author={Kirillov, Alexander and Mintun, Eric and Ravi, Nikhila and Mao, Hanzi and Rolland, Chloe and Gustafson, Laura and Xiao, Tete and Whitehead, Spencer and Berg, Alexander C and Lo, Wan-Yen and others},
  booktitle={Proceedings of the IEEE/CVF international conference on computer vision},
  pages={4015--4026},
  year={2023}
}

@inproceedings{Gaussian-flow2024CVPR,
  title={Gaussian-flow: 4d reconstruction with dynamic 3d gaussian particle},
  author={Lin, Youtian and Dai, Zuozhuo and Zhu, Siyu and Yao, Yao},
  booktitle={Proceedings of the IEEE/CVF Conference on Computer Vision and Pattern Recognition},
  pages={21136--21145},
  year={2024}
}

@inproceedings{RAFT2020ECCV,
  title={Raft: Recurrent all-pairs field transforms for optical flow},
  author={Teed, Zachary and Deng, Jia},
  booktitle={Computer Vision--ECCV 2020: 16th European Conference, Glasgow, UK, August 23--28, 2020, Proceedings, Part II 16},
  pages={402--419},
  year={2020},
  organization={Springer}
}

@article{li2026ipibench,
  title={IPIBench: Evaluating Interactive Proactive Intelligence of MLLMs under Continuous Streams},
  author={Li, Jinzhao and Chen, Yinuo and Song, Wenxuan and Lei, Yijia and Zhang, Yichi and Yan, Honglei and Pan, Panwang and Liu, Miao},
  journal={arXiv preprint arXiv:2605.27074},
  year={2026}
}

@InProceedings{Li_2026_CVPR,
    author    = {Li, Jinzhao and Chen, Yinuo and Piao, Dongxu and Pan, Panwang and Yu, Yifan and Wang, Dong and Yan, Honglei and Yue, Liang and Wang, Shaofei and Chen, Yixin and Huang, Siyuan and Liu, Miao},
    title     = {EgoProx: Evaluating MLLMs on Egocentric 3D Proximity Reasoning Across a Cognitive Hierarchy},
    booktitle = {Proceedings of the IEEE/CVF Conference on Computer Vision and Pattern Recognition (CVPR)},
    month     = {June},
    year      = {2026},
    pages     = {23751-23762}
}

@article{chen2026learning,
  title={Learning Multimodal Embeddings with Evidence-Aligned Readout},
  author={Chen, Zirong and Ye, Fuda and Du, Enjun and Pu, Junfu and Wang, Xinlei and Zuo, Xinyu and Duan, Lisheng and Liang, Haijin and Ma, Jin and Wang, Jiachuan and Zhang, Yongqi},
  journal={arXiv preprint arXiv:2609.33659},
  year={2026}
}

@article{chen2026snapbench,
  title={SnapBench: Benchmarking Snap-and-Ask Multimodal Retrieval for Mobile Interactions},
  author={Chen, Zirong and Ye, Fuda and Zhang, Kuan and Du, Enjun and Pu, Junfu and Wang, Xinlei and Zuo, Xinyu and Duan, Lisheng and Ma, Jin and Zhang, Yongqi},
  journal={arXiv preprint arXiv:2608.29607},
  year={2026}
}
}


\end{document}